\documentclass{article} %
\usepackage{uxbench}

\usepackage{booktabs}
\usepackage{graphicx}
\usepackage{enumitem}
\usepackage{wrapfig}
\usepackage{algorithm}
\usepackage{algpseudocode}
\usepackage{wrapfig}
\usepackage{float}
\usepackage{microtype}
\usepackage{amsmath}
\usepackage{amssymb}
\usepackage{colortbl}
\usepackage[utf8]{inputenc}
\definecolor{lightgray}{rgb}{0.9,0.9,0.9}
\usepackage{caption}
\usepackage{subcaption}
\usepackage{xcolor}
\usepackage{setspace}
\usepackage{url}
\usepackage{multirow}
\usepackage{colortbl}
\usepackage{tabularx}
\usepackage{blindtext}
\usepackage{pgfplots}
\pgfplotsset{compat=1.18} 
\usepackage{tikz}
\usetikzlibrary{er,positioning,bayesnet}
\usepackage{makecell}
\usepackage{tipa}
\usepackage{siunitx}
\usepackage{nicefrac}
\usepackage{tocloft}
\usepackage{listings}
\usepackage[raster,skins]{tcolorbox} %
\usepackage{xltabular}
\usepackage{adjustbox}
\usepackage{xurl}
\usepackage{array}

\makeatother

\usepackage{graphicx}

\usepackage{amsmath, amssymb}
\usepackage{booktabs}
\usepackage{multirow}

\usepackage{algorithm}
\usepackage{algpseudocode}
\usepackage{pifont} 

\usepackage{bbm}
\usepackage{tikz}
\newlength{\barwidth}
\usepackage{lipsum}

\usepackage{hyperref}
\usepackage{CJKutf8}

\title{Cross-Domain Hybrid OPD for Generalizable Search Agents}

\author{
\bfseries Yuanbao Team, Tencent \quad Shanghai Innovation Institute
\thanks{Author contributions listed at the end of the paper (\S \ref{sec: author list}).}
\thanks{Correspondence to: hongzhanchen@sii.edu.cn, liuxiaoyuu66@gmail.com.}
}

\begin{document}

\maketitle

\begin{abstract}

Recent advances in Reinforcement Learning (RL) have substantially improved the capabilities of autonomous search agents, enabling sophisticated planning, and iterative retrieval over dynamic information sources. However, optimizing language models for specialized search behaviors often incurs an alignment tax, where gains in search performance come at the expense of general-purpose capabilities, limiting their effectiveness as universal assistants.
In this technical report, we present the training framework behind the Yuanbao search agent, designed to achieve search specialization without sacrificing general intelligence. Built upon the Hunyuan3 architecture, our framework combines agentic reinforcement learning for autonomous search with a cross-domain expert On-Policy Distillation (OPD) pipeline. 
Experts specializing in complementary general-purpose domains are distilled into the search-specialized student, restoring and further enhancing its broad capabilities.
Rather than treating specialization and general capability as competing objectives, our hybrid training strategy jointly optimizes both, effectively mitigating the alignment tax. Extensive experiments demonstrate that the resulting model achieves competitive search performance while consistently improving its general-purpose capabilities, providing a favorable balance between specialized execution and broad generalization in real-world search scenarios.
\end{abstract}

\begin{figure*}[!h]
    \centering
    \includegraphics[width=0.98\linewidth]{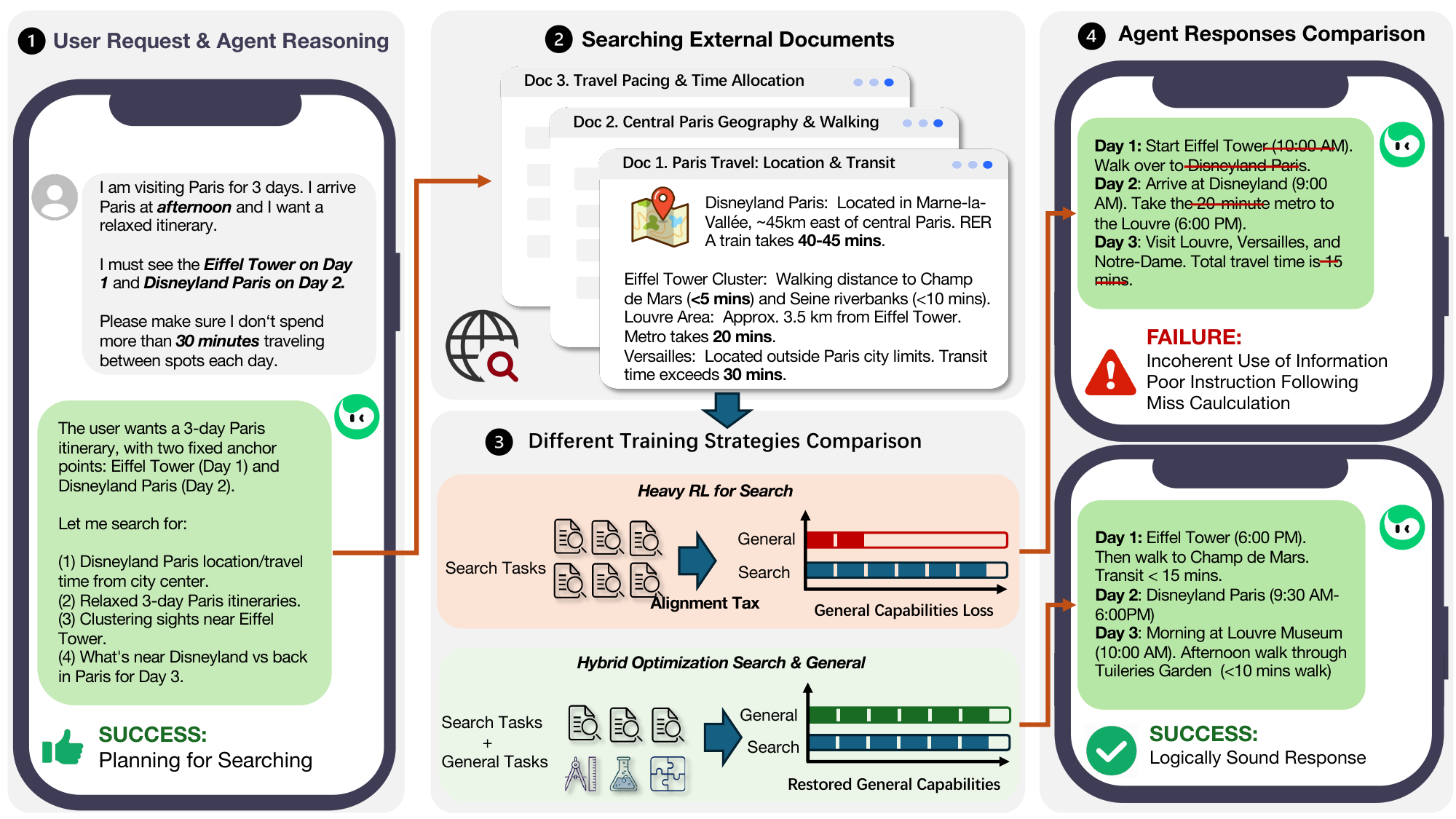}
    \vspace{-0.5em}
\caption{A representative real-world use case of the Yuanbao search agent, illustrating the practical impact of the alignment tax. Search-only RL improves search planning but may weaken general reasoning and instruction following, whereas our hybrid optimization preserves search proficiency while restoring general-purpose capabilities.}
\label{fig:introduction}
\end{figure*}

\section{Introduction}

Large language models (LLMs) are rapidly reshaping information retrieval, driving a shift from conventional search systems toward autonomous search agents capable of planning \citep{jin2025search-r1,jiang2025retrieve}, using tools \citep{li2026webthinker,lumer2025scalemcp}, and multi-step reasoning \citep{zheng2025deepresearcher,wu2025agentic}. 
Unlike conventional retrieval-augmented generation (RAG) \citep{wang2023self,xu2023recomp}, which typically follow a fixed retrieve-then-generate pipeline, modern search agents actively interact with external search engines, iteratively refine search strategies and synthesize information from multiple sources. Reinforcement learning (RL) has emerged as a key paradigm for training such agents, enabling language models to learn effective search policies beyond what can be acquired through supervised fine-tuning alone. As a result, recent search agents have demonstrated remarkable improvements on complex knowledge-intensive tasks requiring long-horizon reasoning and dynamic information acquisition.

Despite these advances, an important challenge remains insufficiently addressed: extensive search-oriented RL can push models toward specialization. While such optimization substantially improves performance on search-intensive tasks, it may weaken the broad capabilities expected of a general-purpose assistant, as illustrated in Figure 1. We view this trade-off as a form of \textit{alignment tax} \citep{ouyang2022training}, manifested in weaker instruction following, reduced conversational quality, and degraded performance on general reasoning and generation tasks. This issue is particularly consequential for real-world AI assistants, where search constitutes only one part of a much broader interaction space. Users expect a search agent not only to retrieve information effectively, but also to write documents, solve problems, perform general reasoning, and engage in natural conversations. The central challenge, therefore, is to improve search specialization without compromising general-purpose intelligence.


To address this challenge, we introduce a hybrid training framework for the Yuanbao Search Agent, built on the Hunyuan3 \citep{hunyuan3preview2026} foundation model. The framework comprises two complementary optimization stages. First, agentic RL equips the model with advanced search behaviors, including autonomous planning, iterative retrieval, and evidence-based reasoning in dynamic environments. Second, a multi-domain expert On-Policy Distillation (OPD) pipeline transfers knowledge from diverse expert models to the search-specialized model, strengthening its general-purpose capabilities while preserving the search policies acquired through RL.

The key insight underlying our framework is that search specialization and general intelligence need not constitute a zero-sum trade-off. Rather than merely repairing the capability degradation introduced by search-oriented RL, our training pipeline treats specialization and generalization as complementary objectives. By integrating search-oriented optimization with multi-domain expert distillation, the framework incorporates diverse expert knowledge without overwriting the search policies acquired through RL. This enables the model to improve as both a search agent and a general-purpose assistant, effectively mitigating the alignment tax associated with search-oriented optimization without sacrificing general capabilities.


Empirical evaluations show that the proposed hybrid training framework achieves a strong balance between specialization and generalization. The resulting Yuanbao search agent attains highly competitive performance on search benchmarks while improving upon the Hunyuan3 foundation model across a broad range of general-purpose evaluations. These results demonstrate that search specialization can be achieved without sacrificing general-purpose capabilities. The contributions of this technical report are summarized as follows:


\begin{enumerate}
    \item We formulate the joint optimization of search specialization and general-purpose capability as a central objective for training universal search agents.
    \item We introduce a hybrid training framework for the Yuanbao search agent that combines agentic reinforcement learning with multi-domain expert On-Policy Distillation on the Hunyuan3 backbone.
    \item We demonstrate that the proposed framework effectively mitigates the alignment tax, preserving highly competitive search performance while improving general-purpose capabilities across diverse benchmarks.
\end{enumerate}


\section{Related Work}

\subsection{LLM-Based Search Agents}

Large language models have extended retrieval-augmented generation (RAG) from relatively fixed retrieval pipelines toward agentic search systems capable of planning, iterative retrieval, and multi-step reasoning. Unlike conventional RAG systems that follow predefined retrieval procedures, modern search agents actively interact with external environments, adapt their search strategies over multiple iterations, and synthesize evidence from diverse sources to solve complex knowledge-intensive tasks.


Early studies primarily focused on agent architectures and workflow design. MindSearch \citep{chen2025mindsearch}, for example, organizes complex information seeking into a dynamic graph involving query decomposition, retrieval, and reasoning. Commercial systems such as OpenAI Deep Research \citep{openai2025deepresearch} and Gemini Deep Research \citep{gemini2025deepresearch} further demonstrate the effectiveness of long-horizon web exploration, enabling agents to autonomously collect evidence and generate comprehensive research reports.


More recent work has increasingly integrated reasoning and search through reinforcement learning. Search-R1 \citep{jin2025search-r1} and R1-Searcher \citep{song2025r1-searcher} formulate search as a sequential decision-making process and optimize how models invoke and interact with external search tools. DeepResearcher \citep{zheng2025deepresearcher} further extends reinforcement learning to realistic web environments, allowing agents to learn search trajectories through direct environmental interaction. Meanwhile, foundation models such as DeepSeek-V4 \citep{xu2026deepseek} and GLM-5 \citep{zeng2026glm} have incorporated stronger tool-use and long-horizon reasoning capabilities into their post-training pipelines. Despite these advances, existing approaches primarily optimize search-specific behaviors. Intensive search-oriented reinforcement learning may bias models toward specialized tool-use patterns and compromise their general-purpose capabilities, resulting in a specialization cost that we refer to as the \emph{alignment tax}. Improving agentic search capabilities without sacrificing general-purpose performance therefore remains an important challenge.


\subsection{On-Policy Distillation}

On-policy distillation (OPD) has emerged as an effective post-training approach for transferring capabilities between large language models. Unlike conventional offline knowledge distillation based on fixed teacher-generated or reference trajectories, OPD samples trajectories from the student’s current policy and uses a stronger teacher to provide supervision over the resulting states. By aligning supervision with the student’s evolving on-policy distribution, OPD provides guidance on the states that the student is likely to encounter during autoregressive generation, thereby reducing the mismatch between distillation and inference and enabling more effective capability transfer.


Early works such as Generalized Knowledge Distillation (GKD) \citep{agarwal2024gkd} and MiniLLM \citep{gu2024minillm} established the foundations of OPD by incorporating student-generated trajectories into the distillation process, demonstrating improvements in distribution alignment and generation quality
More recently, OPD has become an essential component of industrial LLM post-training pipelines. Models including Qwen3~\citep{yang2025qwen3}, DeepSeek-V4~\citep{xu2026deepseek}, and GLM-5~\citep{zeng2026glm} all adopt variants of on-policy distillation for capability transfer, expert consolidation, or capability retention, highlighting its value as a complement to supervised fine-tuning and reinforcement learning.

Although reinforcement learning has been widely used to improve agentic search and OPD has shown promise in transferring and preserving general model capabilities, these two research directions have largely evolved independently. Existing search-agent training pipelines primarily optimize specialized search behaviors, whereas OPD is primarily employed to improve foundation-model capabilities without explicitly considering specialized agent behaviors. In contrast, our work jointly optimizes search-oriented reinforcement learning and multi-domain expert OPD within a unified training framework. This hybrid approach mitigates the alignment tax associated with search specialization, enabling the model to acquire stronger search capabilities while preserving and improving its general-purpose performance.

\begin{table}[t]
\centering
\small
\caption{Summary of the domain-specific training corpora used to train expert teachers. Each corpus combines public benchmarks with internally curated or synthesized data. Difficulty levels are determined by empirical Pass@$k$ statistics.}
\label{tab:domain_data}
\begin{tabular}{p{1.6cm}p{4.3cm}p{4.2cm}p{3.0cm}}
\toprule
\textbf{Domain} & \textbf{Training Sources} & \textbf{Internal Data} & \textbf{Difficulty Levels} \\
\midrule
Mathematics &
MATH, GSM8K, AIME, AMC, AQuA-RAT &
Putnam-/IMO-style problems &
Easy/Medium/Hard \\
\addlinespace

Coding &
LiveCodeBench, CodeContests, APPS, Codeforces, Aider &
Synthesized programming &
Easy/Medium/Hard \\
\addlinespace

Logical Reasoning &
ARC-AGI, SynLogic, BBH &
Board games, puzzle games, compositional and knowledge reasoning &
Medium/Hard \\
\addlinespace

Science &
GPQA, SciQ, SciBench &
Scientific reasoning datasets &
Medium/Hard \\
\bottomrule
\end{tabular}
\end{table}

\section{Data and Environment}

\subsection{Domain Data Construction}

To train domain-specialized expert teachers, we construct four large-scale training corpora covering mathematics, coding, logical reasoning, and science. As summarized in Table~\ref{tab:domain_data}, each corpus combines widely adopted public benchmarks including MATH \citep{hendrycks2021measuring}, GSM8K \citep{cobbe2021training}, AQuA-RAT \citep{ling-etal-2017-program}, APPS \citep{hendrycks2021apps}, LiveCodeBench \citep{jain2025livecodebench}, SynLogic \citep{liu2026synlogic}, BBH \citep{suzgun2023challenging}, SciBench \citep{wang2023scibench}, SciQ \citep{welbl2017crowdsourcing} and GPQA \citep{rein2023gpqa}, with internally curated or synthesized data. Public datasets, after simple filtering, provide standardized evaluation distributions and diverse reasoning tasks, while internally constructed data substantially enrich problem diversity and long-tail reasoning scenarios that are underrepresented in existing benchmarks. Together, these complementary data sources enable each expert teacher to acquire robust and domain-specific reasoning capabilities.

To facilitate efficient reinforcement learning, we further organize each domain corpus according to problem difficulty. Specifically, the difficulty of each training instance is estimated using the empirical success rate of a reference policy, measured by the proportion of successful solutions among multiple sampled attempts. Mathematics and coding tasks are partitioned into three difficulty levels: \emph{easy} (success rate $>0.9$), \emph{medium} ($0.3 \leq$ success rate $\leq 0.9$), and \emph{hard} (success rate $<0.3$). In contrast, logical reasoning and science datasets exhibit relatively concentrated difficulty distributions and are therefore divided into two levels: \emph{medium} (success rate $>0.3$) and \emph{hard} (success rate $\leq0.3$). This difficulty-aware organization serves as the basis of the curriculum learning strategy described in Section~\ref{sec:stage-II}, enabling the training process to prioritize informative samples and progressively improve the reasoning capabilities of each expert teacher.

\subsection{Agent Search Environment}

Training an autonomous search agent requires an environment that closely resembles real-world information retrieval. To this end, we construct an interactive search environment that integrates external retrieval tools with a structured interaction protocol. Rather than directly generating final responses, the agent can iteratively retrieve external information and produce grounded responses. 

\paragraph{Search Tools} The agent is equipped with three retrieval tools: \texttt{web\_search} for full-text document retrieval, \texttt{image\_search} for image retrieval, and \texttt{video\_search} for semantic video retrieval. All tools expose a unified JSON-based interface that specifies the required input arguments and return formats. The system prompt specification is summarized in Table~\ref{tab:system_prompt} in the Appendix.

\paragraph{Tool Invocation Protocol} Instead of directly accessing external services, the model communicates with the environment through a structured XML-style protocol. When external information is required, the model emits one or more tool calls enclosed within the \texttt{<tool\_calls>} tag, where each tool invocation explicitly specifies the tool name together with its input arguments. The execution engine parses the generated tool calls, invokes the corresponding retrieval services, and appends the returned results to the dialogue history as tool observations. The model subsequently conditions on these observations to determine whether additional retrieval is necessary or to generate the final response. This retrieval--execution--generation loop enables iterative query refinement and multi-hop information acquisition over external knowledge sources.



\subsection{Reward and Verification}
\label{sec:verifier}

Reliable reinforcement learning requires accurate reward estimation across heterogeneous tasks. Since different domains possess distinct evaluation characteristics, we adopt a unified verification framework consisting of rule-based evaluators, executable validators, and learned reward models.

\paragraph{Rule-based Verification} For tasks with deterministic ground-truth answers, including many mathematical, logical, and scientific problems, correctness is computed through exact matching.

\paragraph{Reward Models} For problems whose solutions admit multiple valid expressions or free-form responses, we train domain-specific reward models. For mathematics, logic, and science, binary reward models are trained using binary cross-entropy to predict whether the generated answer is semantically equivalent to the reference solution. For search tasks, we instead train a Bradley–Terry preference model using pairwise preference data. The model assigns a scalar reward to each response, with score differences determining the predicted preference probability between response pairs.

\paragraph{Code Execution} Coding tasks are evaluated within a sandbox. Generated programs are executed against predefined test cases, and reward is computed via Pass@k.






\begin{figure*}[t]
    \centering
    \includegraphics[width=0.98\linewidth]{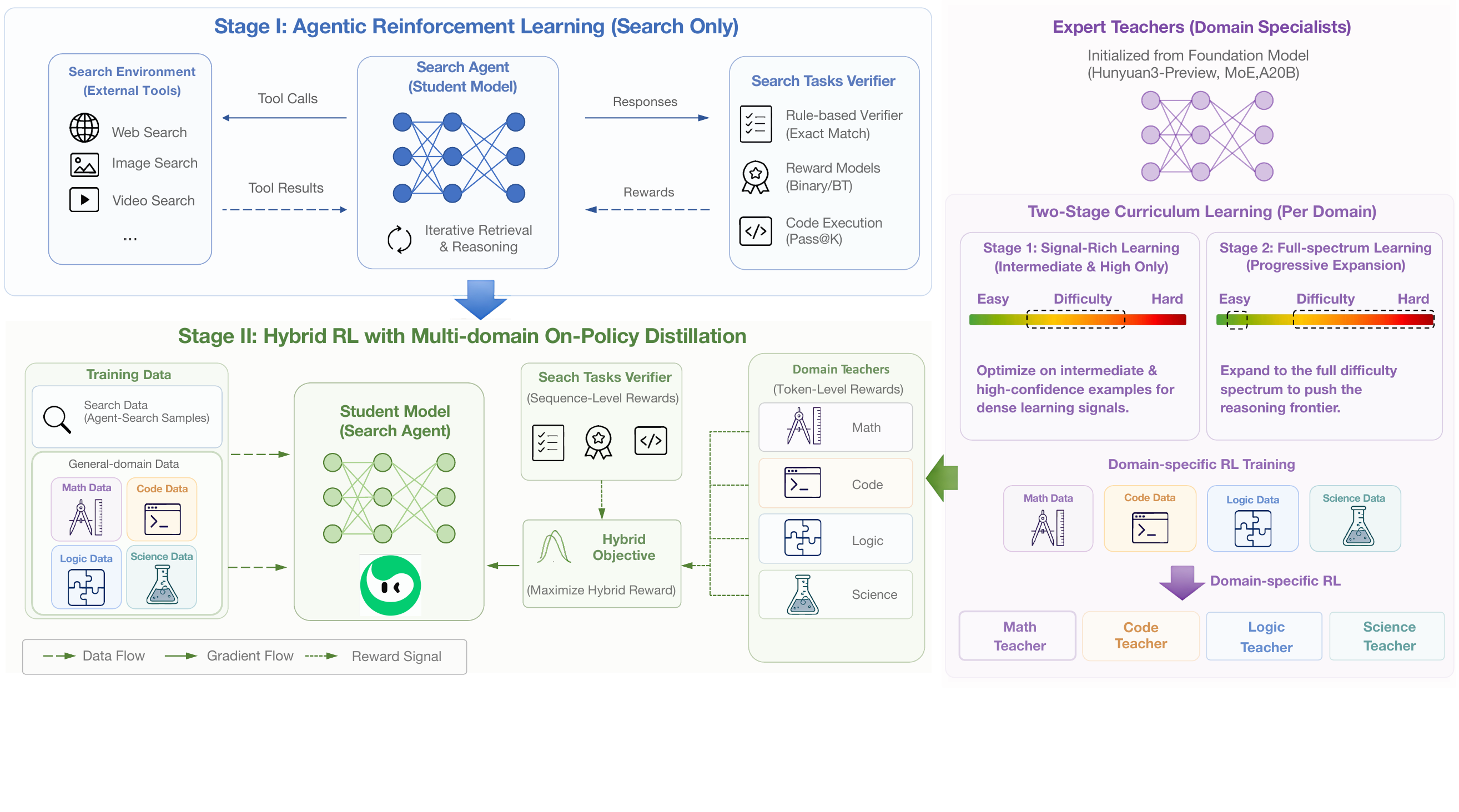}
    \vspace{-15mm}
\caption{Overview of the proposed cross-domain hybrid OPD framework. Stage I trains the search agent using RL on agent search tasks, acquiring autonomous search behaviors. In parallel, domain-specific expert teachers are optimized through a signal-density-driven curriculum. Stage II jointly optimizes search RL and multi-domain OPD, enabling the model to preserving search capabilities while recovering general-purpose performance.}
\label{fig:overview_of_pipeline}
\end{figure*}

\begin{figure*}[t]
    \centering
    \includegraphics[width=0.99\linewidth]{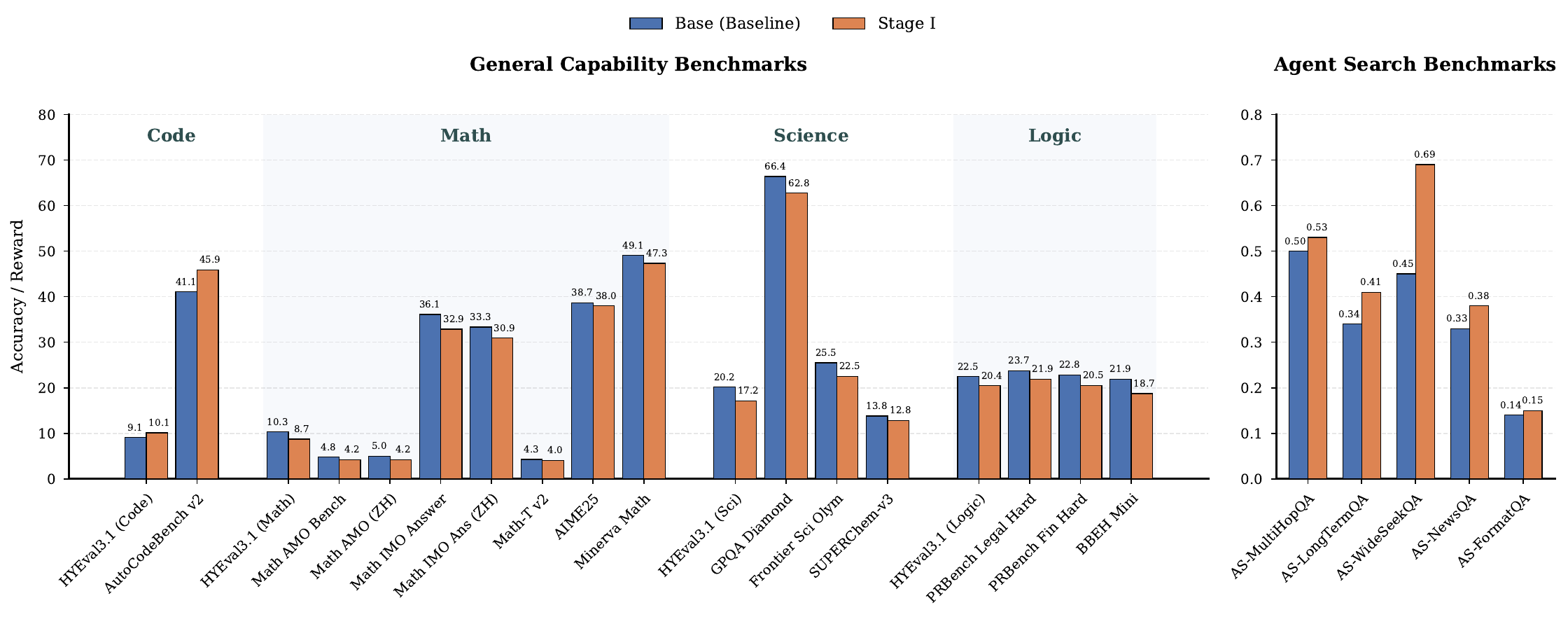}
    \vspace{-0.5em}
\caption{Preliminary experiments on the Hunyuan3 (A3B). Accuracy is reported on representative general capability benchmarks, while average reward is reported on agent search benchmarks. Optimizing the model exclusively on agent search tasks during Stage I consistently improves search performance but degrades performance on most general capability benchmarks, particularly in mathematics, science, and logical reasoning. This trade-off illustrates the \emph{alignment tax} associated with search-oriented reinforcement learning and motivates the hybrid optimization strategy introduced in Stage II.}
\label{fig:model-comparison}
\end{figure*}

\section{Cross-Domain Hybrid OPD}

\subsection{Overview}


We present Hybrid-OPD, a training framework that jointly optimizes agentic search capabilities and general-purpose performance by integrating search-oriented reinforcement learning with multi-domain on-policy distillation. Figure \ref{fig:overview_of_pipeline} illustrates the overall training pipeline.

\subsection{Stage I: Agentic Reinforcement Learning}

The first stage of training aims to endow the foundation model with autonomous search capabilities. Starting from the Hunyuan3 foundation model, we optimize the policy exclusively on agent search trajectories using GRPO~\citep{shao2024deepseekmath}. The reward is obtained from the task-specific verifier introduced in Section~\ref{sec:verifier}, including rule-based verifiers, reward models, and executable code evaluation depending on the task type. This encourages the policy to learn effective planning strategies, invoke external tools when necessary, iteratively refine search queries, and synthesize grounded responses from retrieved evidence.

\subsection{Stage II: Hybrid Reinforcement Learning with Multi-domain On-Policy Distillation}
\label{sec:stage-II}


While Stage I significantly improves autonomous search capabilities, we observe that optimizing solely on search tasks incurs an alignment tax, leading to degraded performance on general-purpose reasoning tasks. For a controlled and compute-efficient analysis, we examine this phenomenon on the Hunyuan3 (A3B) variant. As shown in Figure 3 \ref{fig:model-comparison}, search performance improves consistently after Stage I, whereas performance declines on most general capability benchmarks. To preserve the search capabilities learned in Stage I while recovering broad-domain competence, we introduce a second training stage that jointly optimizes search reinforcement learning and multi-domain on-policy distillation (OPD).

\paragraph{Expert Teachers.}
Before hybrid training, we independently train four expert teacher models specializing in mathematics, coding, logical reasoning, and science, respectively. We adopt a two-stage curriculum learning strategy. Rather than uniformly sampling from all available training instances, each domain dataset is first partitioned into multiple difficulty levels. The first stage focuses exclusively on intermediate-difficulty examples that exhibit high reward variance, thereby providing dense optimization signals and rapidly establishing robust reasoning patterns. Extremely easy samples are intentionally excluded at this stage because they contribute little useful learning signal, while overly difficult samples that receive nearly zero rewards are postponed until later optimization.
Once the teacher model has acquired stable reasoning behaviors, the second stage gradually expands the curriculum to include the full difficulty spectrum. More challenging examples are assigned higher sampling weights to continuously improve the model's reasoning frontier, while a small proportion of easier examples are retained as capability anchors to prevent performance degradation during long-horizon reinforcement learning. This signal-density-driven curriculum balances optimization efficiency and model robustness, allowing the expert teachers to progressively improve on difficult reasoning tasks.


\paragraph{Hybrid Optimization.}
At each training iteration, we construct a mixed mini-batch
$
\mathcal{B}
=
\mathcal{B}_{\text{search}}
\cup
\mathcal{B}_{\text{general}},
$
consisting of search trajectories and general-domain examples.
Search samples continue to be optimized using the Stage-I GRPO objective. For each general-domain sample $(x,d)$, where $d\in\{\text{Math},\text{Code},\text{Logic},\text{Science}\}$ denotes its domain, the corresponding expert teacher $\pi_{\phi_d}$ produces a target distribution. The student policy is then optimized by minimize the reverse KL divergence,
\begin{equation}
\min_\theta~
\mathbb{E}_{(x,d)\sim\mathcal{B}_{\text{general}}}
\mathbb{E}_{y_t\sim\pi_\theta(\cdot|x,y_{<t})}
\left[\log\frac{\pi_\theta(y_t|x,y_{<t})}{\pi_{\phi_d}(y_t|x,y_{<t})}\right].
\label{eq:opd-objective}
\end{equation}
Since the reverse KL is naturally expressed as an expectation under the student distribution, it admits an unbiased Monte Carlo estimator. Following \cite{schulman2020kl}, we use the estimator $k_1=\log\frac{\pi_\theta(y_t|s,y_{<t})}{\pi_{\phi_d}(y_t|x,y_{<t})}$, which is an unbiased estimator of the reverse KL, and treat its negative value as a token-level supervision signal. This formulation enables the distillation objective to be optimized using the same policy-gradient framework as GRPO:
\begin{equation}
\label{eq:hybrid}
\begin{aligned}
    \mathcal{L}&=\mathbb{E}_{x\sim\mathcal{B}_\text{general}\cup\mathcal{B}_\text{search},\{y\}^G_{i=1}\sim\pi_\text{old}(\cdot|x)} \\
    & \left[\frac{1}{G}\sum^G_{i=1}\frac{1}{|y_i|}\sum^{|y_i|}_{t=1}\left\{\min\left[\frac{\pi_\theta(y_{i,t}|x,y_{i,<t})}{\pi_\text{old}(y_{i,t}|x,y_{i,<t})}A_{i,t},\text{clip}\left(\frac{\pi_\theta(y_{i,t}|x,y_{i,<t})}{\pi_\text{old}(y_{i,t}|x,y_{i,<t})},1-\epsilon,1+\epsilon\right)A_{i,t}\right]\right\}\right],
\end{aligned}
\end{equation}
where:
\begin{equation}
    A_{i,t}=\begin{cases}
        \frac{R_i-\text{mean}(\{R\}^G_{i=1})}{\text{std}(\{R\}^G_{i=1})},&\text{if } x\in\mathcal{B}_\text{search}, \\
        -\text{sg}\left[\log\frac{\pi_\text{old}(y_{i,t}|x,y_{i,<t})}{\pi_{\phi_d}(y_{i,t}|x,y_{i,<t})}\right], & \text{otherwise},
    \end{cases}
\end{equation}
where $d$ denotes the domain label associated with the general-domain sample $x$ and $\text{sg}(\cdot)$ is the stop-gradient function. Algorithm~\ref{alg:hybrid_training} summarizes the training procedure. At each iteration, search trajectories are collected by interacting with the search environment and optimized using GRPO with verifier-based rewards. In parallel, the student generates on-policy responses for general-domain prompts, which are evaluated by the corresponding expert teachers to provide token-level supervision through reverse-KL distillation. The reinforcement learning objective and the OPD objective are jointly optimized according to Eq.~(\ref{eq:hybrid}), enabling the model to simultaneously strengthen autonomous search behavior while recovering broad reasoning competence.

\begin{algorithm}[t]
\caption{Cross-Domain Hybrid RL with OPD}
\label{alg:hybrid_training}
\begin{algorithmic}[1]

\Require
Stage-I search policy $\pi_\theta$;
Expert teachers $\{\pi_{\phi_d}\}_{d\in\mathcal{D}}$;
Search dataset $\mathcal{S}$;
General-domain datasets $\{\mathcal{G}_d\}_{d\in\mathcal{D}}$; Learning rate $\eta$; Group size $G$.

\For{each training iteration}

    \State Construct a mixed mini-batch
    \[
    \mathcal{B}
    =
    \mathcal{B}_{\mathrm{search}}
    \cup
    \mathcal{B}_{\mathrm{general}},
    \]
    \State where
    $\mathcal{B}_{\mathrm{search}}\sim\mathcal{S}$ and
    $\mathcal{B}_{\mathrm{general}}\sim\{\mathcal{G}_d\}_{d\in\mathcal{D}}$.


    \ForAll{$x\in\mathcal{B}_{\mathrm{search}}$}

        \State Generate trajectories $\{y_i\}_{i=1}^{G}$ from $\pi_\theta(\cdot|x)$ and interact with search tools.

        \State Compute reward $R_i$
        using rule-based verifiers, reward models, or execution feedback.

        \State Compute GRPO advantage:
        \[
        A^\text{RL}_{i,t}=
        \frac{R_i-\mathrm{mean}(\{R\}_{i=1}^{G})}
        {\mathrm{std}(\{R\}_{i=1}^{G})}.
        \]
    \EndFor
    \ForAll{$(x,d)\in\mathcal{B}_{\mathrm{general}}$}
        \State Generate trajectories $\{y_i\}_{i=1}^{G}$ from $\pi_\theta(\cdot|x)$.
        
        \State Route the sample to the corresponding expert teacher
        $\pi_{\phi_d}$.

        \State Query teacher server to obtain
        token-level log probabilities
        $\log\pi_{\phi_d}(y_{i,t}|x,y_{i,<t})$.

        \State Compute OPD advantage:
        \[
        A^\text{OPD}_{i,t}=
        -\log
        \frac{
        \pi_\text{old}(y_{i,t}|x,y_{i,<t})
        }{
        \pi_{\phi_d}(y_{i,t}|x,y_{i,<t})
        }.
        \]
    \EndFor

    \State Optimize the unified hybrid objective $\mathcal{L}$ using Equation \ref{eq:hybrid}.
    \State Update policy parameters $\theta \leftarrow \theta-\eta\nabla_\theta\mathcal{L}$.
\EndFor

\State \Return
hybrid-trained search agent $\pi_\theta$.

\end{algorithmic}
\end{algorithm}

\section{Experimental Setup}

\paragraph{Training Data.}

The training corpus consists of two complementary categories: \emph{general reasoning tasks} and \emph{agent search tasks}. General reasoning tasks cover four representative domains, namely mathematics, coding, logical reasoning, and science. Each problem admits either a unique answer or a verifiable solution, allowing rewards to be computed using verifiers. These datasets primarily develop structured reasoning, symbolic manipulation, and algorithmic problem-solving abilities. The agent search corpus is constructed from real-world user queries and consists of six datasets covering diverse scenarios, including general search question answering, search triggering, multi-hop retrieval, reward model training, RLHF preference optimization, and internal production tasks. Rewards are provided by pairwise preference models or category-specific reward models, encouraging the model to learn when to invoke search tools, how to retrieve relevant information, and how to synthesize evidence across multiple search steps.

\paragraph{Models and Training Details.}

We evaluate our framework mainly on Hunyuan3 (A21B) backbone.
For OPD teachers, we train four domain-specialized expert teachers corresponding to mathematics, coding, logical reasoning, and science. Each teacher is initialized from the same A21B base model and optimized exclusively on its corresponding domain using the same GRPO objective and hyperparameters as the student. During hybrid OPD training, each training sample is dynamically routed according to its domain. General reasoning samples are assigned to their corresponding expert teacher, whereas agent-search samples are optimized directly using the standard GRPO objective with rewards provided by the corresponding reward models. Teachers are deployed as independent inference servers that provide reference log-probabilities on demand, avoiding the memory overhead of loading multiple teacher models into the training process. Unless otherwise specified, all experiments use the same optimization configuration. We sample 8 rollouts per prompt, with learning rate 5e-6 and a maximum sequence length of 61,440 tokens 



\paragraph{Evaluation Data.}

We evaluate the proposed framework on two complementary benchmark suites that measure both specialized search capability and general-domain reasoning. The search evaluation consists of five internal benchmarks designed to assess autonomous search behavior from different perspectives. AS-MultiHopQA evaluates multi-hop knowledge retrieval and tool planning through single-turn closed-domain questions requiring multiple web searches. AS-LongTermQA is constructed from real user interactions and human annotations to assess long-context, multi-turn conversational search. AS-WideSeekQA focuses on factual retrieval over knowledge-intensive queries. AS-NewsQA evaluates complex search scenarios involving multi-step reasoning and time-sensitive information, while AS-FormatQA measures instruction-following ability under long and complex prompts with strict formatting requirements. Most search benchmarks are evaluated using LLM-based judgment, either by answer equivalence or checklist verification. To quantify the alignment tax introduced by search-specific optimization, we evaluate general capabilities on a diverse collection of coding, mathematics, science, and logical reasoning benchmarks, including HYEval3.1, AutoCodeBench v2 \citep{chou2025autocodebench}, Math AMO Bench \citep{an2025amo}, Math IMO AnswerBench \citep{luong2025towards}, AIME25 \citep{aops2025aimei}, Minerva Math \citep{lewkowycz2022solving}, GPQA Diamond \citep{rein2023gpqa}, Frontier Science Olympiad \citep{wang2026frontierscience}, SUPERChem-v3 \citep{zhao2025superchem}, PRBench \citep{akyurek-etal-2026-prbench}, and BBEH Mini \citep{kazemi-etal-2025-big}. 

\begin{table}[t]
\centering
\caption{General capability evaluation. Accuracy (\%) on representative coding, mathematics, science, and logical reasoning benchmarks. Stage I performs reinforcement learning exclusively on agent search tasks, while Stage II further introduces multi-domain on-policy distillation. Values in parentheses denote the performance difference relative to the base model.}
\label{tab:main-general}
\small
\begin{tabularx}{\linewidth}{>{\raggedright\arraybackslash}Xccc}
\toprule
\textbf{Benchmark} & \textbf{Base} & \textbf{Stage I} & \textbf{Stage II} \\
\midrule
\midrule

\multicolumn{4}{l}{\textit{Code}}\\
\midrule
HYEval3.1 (Code)                  & 18.51 & 17.86 \textcolor{blue}{(-0.65)} & 19.58 \textcolor{red}{(+1.07)} \\
AutoCodeBench v2                  & 68.94 & 67.74 \textcolor{blue}{(-1.20)} & 74.15 \textcolor{red}{(+5.21)} \\

\addlinespace
\multicolumn{4}{l}{\textit{Mathematics}}\\
\midrule
HYEval3.1 (Math)                  & 18.44 & 16.76 \textcolor{blue}{(-1.68)} & 18.46 \textcolor{red}{(+0.02)} \\
Math AMO Bench                    & 15.38 & 17.75 \textcolor{red}{(+2.37)} & 16.12 \textcolor{red}{(+0.74)} \\
Math AMO Bench (ZH)               & 24.67 & 16.67 \textcolor{blue}{(-8.00)} & 24.00 \textcolor{blue}{(-0.67)} \\
Math IMO AnswerBench              & 50.79 & 48.85 \textcolor{blue}{(-1.94)} & 49.81 \textcolor{blue}{(-0.98)} \\
Math IMO AnswerBench (ZH)         & 50.52 & 48.23 \textcolor{blue}{(-2.29)} & 53.87 \textcolor{red}{(+3.35)} \\
Math-T v2                         & 12.62 & 12.00 \textcolor{blue}{(-0.62)} & 13.41 \textcolor{red}{(+0.79)} \\
AIME25                            & 62.00 & 59.00 \textcolor{blue}{(-3.00)} & 66.33 \textcolor{red}{(+4.33)} \\
Minerva Math                      & 54.23 & 48.35 \textcolor{blue}{(-5.88)} & 52.94 \textcolor{blue}{(-1.29)} \\

\addlinespace
\multicolumn{4}{l}{\textit{Science}}\\
\midrule
HYEval3.1 (Science)               & 33.20 & 34.40 \textcolor{red}{(+1.20)} & 37.83 \textcolor{red}{(+4.63)} \\
GPQA Diamond                      & 73.23 & 69.95 \textcolor{blue}{(-3.28)} & 76.64 \textcolor{red}{(+3.41)} \\
Frontier Science Olympiad         & 55.00 & 52.00 \textcolor{blue}{(-3.00)} & 50.00 \textcolor{blue}{(-5.00)} \\
SUPERChem-v3                      & 20.40 & 22.80 \textcolor{red}{(+2.40)} & 22.60 \textcolor{red}{(+2.20)} \\

\addlinespace
\multicolumn{4}{l}{\textit{Logical Reasoning}}\\
\midrule
HYEval3.1 (Logical Reasoning)     & 37.16 & 33.95 \textcolor{blue}{(-3.21)} & 46.90 \textcolor{red}{(+9.74)} \\
PRBench Legal Hard                & 32.33 & 31.78 \textcolor{blue}{(-0.55)} & 32.79 \textcolor{red}{(+0.46)} \\
PRBench Finance Hard              & 32.42 & 32.18 \textcolor{blue}{(-0.24)} & 33.42 \textcolor{red}{(+1.00)} \\
BBEH Mini                         & 40.22 & 34.60 \textcolor{blue}{(-5.62)} & 47.86 \textcolor{red}{(+7.64)} \\

\bottomrule
\end{tabularx}
\end{table}

\begin{table}[t]
\centering
\caption{Agent search capability evaluation. Average reward on five representative agent search benchmarks. Stage I substantially improves autonomous search behaviors through reinforcement learning, while Stage II further enhances search capability while largely preserving the gains achieved in Stage I and further improving three of the five benchmarks. Values in parentheses denote the improvement over the base model.}
\label{tab:main-agent-search}
\small
\begin{tabularx}{\linewidth}{>{\raggedright\arraybackslash}Xccc}
\toprule
\textbf{Benchmark} & \textbf{Base} & \textbf{Stage I} & \textbf{Stage II} \\
\midrule
\midrule

AS-MultiHopQA                  & 0.5012 & 0.8266 \textcolor{red}{(+0.3254)} & 0.8386 \textcolor{red}{(+0.3374)} \\
AS-LongTermQA                & 0.5093 & 0.6886 \textcolor{red}{(+0.1793)} & 0.6728 \textcolor{red}{(+0.1635)} \\
AS-WideSeekQA               & 0.4267 & 0.7867 \textcolor{red}{(+0.3600)} & 0.8017 \textcolor{red}{(+0.3750)} \\
AS-NewsQA                     & 0.4853 & 0.6250 \textcolor{red}{(+0.1397)} & 0.6107 \textcolor{red}{(+0.1254)} \\
AS-FormatQA                & 0.1922 & 0.2556 \textcolor{red}{(+0.0634)} & 0.2859 \textcolor{red}{(+0.0937)} \\

\bottomrule
\end{tabularx}
\end{table}

\section{Main Results}

We evaluate the proposed framework from two complementary perspectives: general-purpose reasoning across diverse domains and autonomous search capability in realistic search environments. General capability is measured on representative coding, mathematics, science, and logical reasoning benchmarks, while search capability is evaluated using five internal agent search benchmarks covering multi-hop retrieval, long-context interaction, deep information seeking, news-oriented reasoning, and instruction following. Tables~\ref{tab:main-general} and~\ref{tab:main-agent-search} summarize the results.

\subsection{Stage I: Improving Search Capability but Incurring an Alignment Tax}

Stage I optimizes the model exclusively on agent search trajectories using reinforcement learning. As shown in Table~\ref{tab:main-agent-search}, this optimization consistently improves search capability across all five evaluation benchmarks. Compared with the base model, reward increases by more than $0.32$ on AS-MultiHopQA and AS-WideSeekQA, indicating substantially stronger planning, retrieval, and evidence aggregation abilities. Improvements are also observed on AS-LongTermQA, AS-NewsQA, and AS-FormatQA, demonstrating that reinforcement learning effectively equips the model with autonomous search behaviors over diverse interaction scenarios. However, these gains come at the expense of general-purpose capabilities. As shown in Table~\ref{tab:main-general}, most general benchmarks experience performance degradation after Stage I. The decline is particularly pronounced on reasoning-intensive tasks, including Math AMO Bench (ZH) ($-8.00$), Minerva Math ($-5.88$), BBEH Mini ($-5.62$), GPQA Diamond ($-3.28$), HYEval3.1 Logical Reasoning ($-3.21$), and AIME25 ($-3.00$). 
Although several benchmarks exhibit moderate improvements, the overall results reveal a clear trade-off between search specialization and broad reasoning performance. Search-only optimization strengthens agentic search behaviors but degrades performance on most general capability benchmarks, providing empirical evidence of the alignment tax induced by search-oriented reinforcement learning.

\subsection{Stage II: Mitigating the Alignment Tax while Preserving Search Capability}

Stage II jointly optimizes search reinforcement learning and multi-domain on-policy distillation. Compared with Stage I, the hybrid objective restores performance on most general capability benchmarks. The largest gains are observed on logical reasoning benchmarks, where HYEval3.1 Logical Reasoning improves by $12.95$ points relative to Stage I and BBEH Mini improves by $13.26$ points. Coding capability also benefits substantially, with AutoCodeBench v2 improving from $67.74$ to $74.15$, exceeding the base model by more than five percentage points. The recovery is equally evident on mathematical reasoning. AIME25 increases from $59.00$ to $66.33$, while Math AMO Bench (ZH) recovers nearly all of the performance lost during Stage I. Furthermore, Math IMO AnswerBench (ZH) surpasses the base model by $3.35$ points. Similar trends are observed in science benchmarks, where GPQA Diamond improves by $6.69$ points over Stage I and HYEval3.1 Science reaches the highest accuracy among all three models. Although a few extremely challenging benchmarks, such as Frontier Science Olympiad and Minerva Math, remain slightly below the base model, the overall results demonstrate that the proposed hybrid optimization effectively mitigates the capability degradation caused by search reinforcement learning. Importantly, the recovery of general capability does not compromise search performance. As shown in Table~\ref{tab:main-agent-search}, Stage II further improves AS-MultiHopQA, AS-WideSeekQA, and AS-FormatQA beyond the already strong Stage-I model, while only introducing minor decreases on AS-LongTermQA and AS-NewsQA. Across all benchmarks, the search capability remains substantially stronger than that of the base model, indicating that the knowledge injected through expert distillation complements rather than overwrites the search policy learned through RL.

\begin{table}[t]
\centering
\caption{Ablation study of the proposed framework on general capability benchmarks. 
\textbf{Hybrid Optimization} denotes our full method. 
\textbf{Mix RL} jointly optimizes search and general data with GRPO, 
\textbf{Single General Teacher} replaces multiple domain-specific teachers with a single general teacher, and 
\textbf{w/o Curriculum} removes curriculum learning during teacher training. 
The best results are shown in \textbf{bold}, while the second-best results are \underline{underlined}.}
\label{tab:ablation-general}
\small
\begin{tabularx}{\linewidth}{>{\raggedright\arraybackslash}Xccc>{\columncolor{blue!5}}c}
\toprule
\textbf{Benchmark} & \textbf{Mix} & \textbf{Single} & \textbf{w/o} & \textbf{Hybrid} \\
                   & \textbf{RL} & \textbf{General Teacher} & \textbf{Curriculum} & \textbf{Optimization} \\
\midrule
\midrule

\multicolumn{5}{l}{\textit{Code}}\\
\midrule
HYEval3.1 (Code)                  & \underline{19.32} & 17.94 & 18.83 & \textbf{19.58} \\
AutoCodeBench v2                  & 71.34 & 69.94 & \underline{71.84} & \textbf{74.15} \\

\addlinespace
\multicolumn{5}{l}{\textit{Mathematics}}\\
\midrule
HYEval3.1 (Math)                  & \underline{16.99} & 16.85 & 16.94 & \textbf{18.46} \\
Math AMO Bench                    & 12.50 & 12.62 & \underline{15.25} & \textbf{16.12} \\
Math AMO Bench (ZH)               & 13.38 & 14.37 & \underline{18.67} & \textbf{24.00} \\
Math IMO AnswerBench              & 47.50 & 48.56 & \underline{49.10} & \textbf{49.81} \\
Math IMO AnswerBench (ZH)         & 47.19 & 45.12 & \underline{49.02} & \textbf{53.87} \\
Math-T v2                         & 11.49 & 11.73 & \underline{11.90} & \textbf{13.41} \\
AIME25                            & \underline{66.33} & \textbf{70.00} & \underline{66.33} & \underline{66.33} \\
Minerva Math                      & \textbf{53.68} & 50.74 & \underline{53.49} & 52.94 \\

\addlinespace
\multicolumn{5}{l}{\textit{Science}}\\
\midrule
HYEval3.1 (Science)               & 33.68 & 33.30 & \underline{34.31} & \textbf{37.83} \\
GPQA Diamond                      & 73.61 & \underline{75.51} & \textbf{76.64} & \textbf{76.64} \\
Frontier Science Olympiad         & 45.00 & \underline{48.00} & 46.75 & \textbf{50.00} \\
SUPERChem-v3                      & 21.04 & \underline{23.00} & \textbf{23.20} & 22.60 \\

\addlinespace
\multicolumn{5}{l}{\textit{Logical Reasoning}}\\
\midrule
HYEval3.1 (Logical Reasoning)     & 36.31 & \underline{40.35} & 36.64 & \textbf{46.90} \\
PRBench Legal Hard                & 29.28 & \underline{29.31} & 28.71 & \textbf{32.79} \\
PRBench Finance Hard              & \underline{28.45} & 28.19 & 28.41 & \textbf{33.42} \\
BBEH Mini                         & 35.22 & 39.60 & \underline{40.65} & \textbf{47.86} \\

\bottomrule
\end{tabularx}
\end{table}

\begin{table}[t]
\centering
\caption{Ablation study of curriculum learning on general validation sets.
We compare teachers trained with and without curriculum learning across different difficulty levels. 
The medium and hard levels are further split into L1 and L2, where L2 corresponds to more challenging problems. Values in parentheses denote performance differences relative to the curriculum-trained teacher.}
\label{tab:ablation-curriculum-math}
\small
\setlength{\tabcolsep}{4.5pt}
\begin{tabular}{lccccc|c}
\toprule
\textbf{Teacher Method} 
& \textbf{Easy} 
& \textbf{Medium-L1} 
& \textbf{Medium-L2} 
& \textbf{Hard-L1} 
& \textbf{Hard-L2}
& \textbf{Average}
\\
\midrule

w/ Curriculum
& 97.83 
& {75.26}
& {62.54}
& {36.43}
& {36.42} 
& 61.70\\

w/o Curriculum 
& 98.00 \textcolor{red}{(+0.17)}
& 76.98 \textcolor{red}{(+1.72)}
& 59.79 \textcolor{blue}{(-2.75)}
& 38.15 \textcolor{red}{(+1.72)}
& 30.93 \textcolor{blue}{(-5.49)} 
& 60.77 \textcolor{blue}{(-0.93)}\\

\bottomrule
\end{tabular}
\end{table}

\section{Ablation Study}

To investigate how each component contributes to mitigating the alignment tax induced by search-oriented training, we conduct ablation studies on general capability benchmarks. Results are presented in Table~\ref{tab:ablation-general}.

\subsection{Effect of Cross-Domain On-Policy Distillation}

To evaluate the contribution of cross-domain OPD in Stage II, we compare Hybrid Optimization with a variant that replaces the OPD objective with outcome-based GRPO over all training samples ({Mix RL}). As shown in Table~\ref{tab:ablation-general}, removing OPD results in substantial degradation across general capability benchmarks, demonstrating the difficulty of recovering general intelligence through RL optimization alone. The degradation is particularly pronounced on reasoning-intensive tasks. For instance, {Mix RL} reduces the accuracy on HYEval3.1 (Logical Reasoning) from 46.90\% to 36.31\% and on BBEH Mini from 47.86\% to 35.22\%. Similar trends are observed in mathematical reasoning, where Hybrid Optimization achieves 24.00\% and 53.87\% on Math AMO Bench (ZH) and Math IMO AnswerBench (ZH), respectively, substantially outperforming {Mix RL} (13.38\% and 47.19\%). These results indicate that outcome-based RL alone cannot effectively mitigate the alignment tax introduced by search-oriented training. By contrast, OPD provides dense token-level supervision, allowing the model to retain and further improve its general reasoning capabilities.

\subsection{Effect of Multi Expert Teachers}

We further investigate the importance of domain-specific expert teachers by replacing the four expert teachers with a single general teacher trained across all domains (Single General Teacher). Compared with the full framework, this variant underperforms the full framework on most benchmarks. The performance degradation is particularly noticeable on tasks requiring specialized reasoning patterns. For instance, the single-teacher variant achieves only 40.35\% on HYEval3.1 (Logical Reasoning), compared with 46.90\% obtained by Hybrid Optimization. Similarly, on AutoCodeBench v2, the single-teacher variant achieves 69.94\%, while the proposed multi-teacher framework improves the score to 74.15\%. These results indicate that a unified teacher trained across heterogeneous domains cannot provide equally effective supervision for diverse reasoning capabilities. By separating expert teachers for different domains, the proposed framework enables each teacher to capture domain-specific reasoning trajectories and provide more accurate supervision during OPD. The consistent improvements across different domains verify that multi-domain expert routing is essential for effective capability recovery and generalization.

\subsection{Effect of Curriculum Learning}

Curriculum learning plays a critical role in improving the quality of expert teachers, which directly affects the effectiveness of subsequent OPD optimization. To investigate its impact, we perform an ablation study by removing curriculum learning during teacher training and comparing the resulting teacher models across different difficulty levels. As shown in Table~\ref{tab:ablation-curriculum-math}, removing curriculum learning leads to a substantial degradation on the most challenging levels, achieving only 30.93\% accuracy on Hard-L2 problems, compared with 36.42\% for the curriculum-trained teacher. Similar performance drops are observed on the average accuracy. Instead of directly exposing the model to extremely challenging instances, the curriculum strategy gradually builds reliable reasoning behaviors from intermediate-level examples before expanding to harder tasks. As a result, the expert teachers develop stronger capabilities on challenging reasoning problems while maintaining comparable performance on easier examples. The improved teacher quality further benefits the downstream hybrid optimization. As shown in Table~\ref{tab:ablation-general}, removing curriculum learning during teacher training (w/o Curriculum) consistently reduces the performance of the final search agent across multiple general capability benchmarks. For example, the full Hybrid Optimization method improves over the w/o Curriculum variant on most benchmarks. These results demonstrate that curriculum-trained teachers provide more informative and reliable supervision during OPD, allowing the student model to absorb stronger domain-specific reasoning capabilities. 



\section{Conclusion}

In this technical report, we present the hybrid training framework behind the Yuanbao search agent, which addresses the trade-off between search specialization and general-purpose capabilities. By jointly optimizing search-oriented reinforcement learning and multi-domain on-policy distillation, our approach enables the model to acquire autonomous search behaviors while recovering and preserving broad-domain abilities. Experiments across agent search and general capability benchmarks demonstrate that the proposed framework mitigates the alignment tax introduced by search-only optimization and achieves strong performance in both settings. These results show that search specialization and general-purpose performance can be jointly optimized, providing a practical pathway toward building more capable and versatile search agents.


\section*{Full Author List}
\label{sec: author list}
Hongzhan Chen$^\dagger$, Xiaoyu Liu$^\dagger$ , Dengming Zhang, Minzhou Huang, Dongliang Xu, Jingcheng Xie, Dongxiang Fang, Bowen Qin, Minsheng Hao, Yaozong Shen, Xiaojun Quan, Mona Zhou (project leader), Haosheng Zou (project leader),  Jeff Chen (project leader)

\bibliography{references}

@article{yang2025qwen3,
  title={Qwen3 technical report},
  author={Yang, An and Li, Anfeng and Yang, Baosong and Zhang, Beichen and Hui, Binyuan and Zheng, Bo and Yu, Bowen and Gao, Chang and Huang, Chengen and Lv, Chenxu and others},
  journal={arXiv preprint arXiv:2505.09388},
  year={2025}
}

@article{zeng2026glm,
  title={Glm-5: from vibe coding to agentic engineering},
  author={Zeng, Aohan and Lv, Xin and Hou, Zhenyu and Du, Zhengxiao and Zheng, Qinkai and Chen, Bin and Yin, Da and Ge, Chendi and Huang, Chenghua and Xie, Chengxing and others},
  journal={arXiv preprint arXiv:2602.15763},
  year={2026}
}

@article{xu2026deepseek,
  title={DeepSeek-V4: Towards Highly Efficient Million-Token Context Intelligence},
  author={Xu, Anyi and Lin, Bangcai and Xue, Bing and Wang, Bingxuan and Xu, Bingzheng and Wu, Bochao and Zhang, Bowei and Lin, Chaofan and Dong, Chen and Ling, Chenchen and others},
  journal={arXiv preprint arXiv:2606.19348},
  year={2026}
}

@article{jin2025search-r1,
  title={Search-r1: Training llms to reason and leverage search engines with reinforcement learning},
  author={Jin, Bowen and Zeng, Hansi and Yue, Zhenrui and Yoon, Jinsung and Arik, Sercan and Wang, Dong and Zamani, Hamed and Han, Jiawei},
  journal={arXiv preprint arXiv:2503.09516},
  year={2025}
}

@inproceedings{zheng2025deepresearcher,
  title={Deepresearcher: Scaling deep research via reinforcement learning in real-world environments},
  author={Zheng, Yuxiang and Fu, Dayuan and Hu, Xiangkun and Cai, Xiaojie and Ye, Lyumanshan and Lu, Pengrui and Liu, Pengfei},
  booktitle={Proceedings of the 2025 Conference on Empirical Methods in Natural Language Processing},
  pages={414--431},
  year={2025}
}

@article{li2026webthinker,
  title={Webthinker: Empowering large reasoning models with deep research capability},
  author={Li, Xiaoxi and Jin, Jiajie and Dong, Guanting and Qian, Hongjin and Wu, Yongkang and Wen, Ji-Rong and Zhu, Yutao and Dou, Zhicheng},
  journal={Advances in Neural Information Processing Systems},
  volume={38},
  pages={120091--120131},
  year={2026}
}

@inproceedings{lumer2025scalemcp,
  title={Scalemcp: Dynamic and auto-synchronizing model context protocol tools for llm agents},
  author={Lumer, Elias and Gulati, Anmol and Subbiah, Vamse Kumar and Basavaraju, Pradeep Honaganahalli and Burke, James A},
  booktitle={International Joint Conference on Computational Intelligence},
  pages={23--42},
  year={2025},
  organization={Springer}
}

@inproceedings{wu2025agentic,
  title={Agentic reasoning: A streamlined framework for enhancing llm reasoning with agentic tools},
  author={Wu, Junde and Zhu, Jiayuan and Liu, Yuyuan and Xu, Min and Jin, Yueming},
  booktitle={Proceedings of the 63rd Annual Meeting of the Association for Computational Linguistics (Volume 1: Long Papers)},
  pages={28489--28503},
  year={2025}
}

@inproceedings{jiang2025retrieve,
  title={Retrieve, summarize, plan: Advancing multi-hop question answering with an iterative approach},
  author={Jiang, Zhouyu and Sun, Mengshu and Liang, Lei and Zhang, Zhiqiang},
  booktitle={Companion Proceedings of the ACM on Web Conference 2025},
  pages={1677--1686},
  year={2025}
}

@article{ouyang2022training,
  title={Training language models to follow instructions with human feedback},
  author={Ouyang, Long and Wu, Jeffrey and Jiang, Xu and Almeida, Diogo and Wainwright, Carroll and Mishkin, Pamela and Zhang, Chong and Agarwal, Sandhini and Slama, Katarina and Ray, Alex and others},
  journal={Advances in neural information processing systems},
  volume={35},
  pages={27730--27744},
  year={2022}
}

@inproceedings{wang2023self,
  title={Self-knowledge guided retrieval augmentation for large language models},
  author={Wang, Yile and Li, Peng and Sun, Maosong and Liu, Yang},
  booktitle={Findings of the Association for Computational Linguistics: EMNLP 2023},
  pages={10303--10315},
  year={2023}
}

@article{xu2023recomp,
  title={Recomp: Improving retrieval-augmented lms with compression and selective augmentation},
  author={Xu, Fangyuan and Shi, Weijia and Choi, Eunsol},
  journal={arXiv preprint arXiv:2310.04408},
  year={2023}
}

@inproceedings{gu2024minillm,
  title={Minillm: Knowledge distillation of large language models},
  author={Gu, Yuxian and Dong, Li and Wei, Furu and Huang, Minlie},
  booktitle={International Conference on Learning Representations},
  volume={2024},
  pages={32694--32717},
  year={2024}
}

@inproceedings{chen2025mindsearch,
  title={Mindsearch: Mimicking human minds elicits deep ai searcher},
  author={Chen, Zehui and Liu, Kuikun and Wang, Qiuchen and Liu, Jiangning and Zhang, Wenwei and Chen, Kai and Zhao, Feng},
  booktitle={International Conference On Learning Representations},
  volume={2025},
  pages={90007--90029},
  year={2025}
}

@online{openai2025deepresearch,
  author = {{OpenAI}},
  title = {Introducing Deep Research},
  year = {2025},
  url = {https://openai.com/index/introducing-deep-research/},
  urldate = {2025-06-26}
}

@online{gemini2025deepresearch,
  author = {{Gemini}},
  title = {Gemini Deep Research},
  year = {2025},
  url = {https://gemini.google/overview/deep-research/},
  urldate = {2025-06-26}
}

@article{shao2024deepseekmath,
  title={Deepseekmath: Pushing the limits of mathematical reasoning in open language models},
  author={Shao, Zhihong and Wang, Peiyi and Zhu, Qihao and Xu, Runxin and Song, Junxiao and Bi, Xiao and Zhang, Haowei and Zhang, Mingchuan and Li, YK and Wu, Yang and others},
  journal={arXiv preprint arXiv:2402.03300},
  year={2024}
}

@article{song2025r1-searcher,
  title={R1-searcher: Incentivizing the search capability in llms via reinforcement learning},
  author={Song, Huatong and Jiang, Jinhao and Min, Yingqian and Chen, Jie and Chen, Zhipeng and Zhao, Wayne Xin and Fang, Lei and Wen, Ji-Rong},
  journal={arXiv preprint arXiv:2503.05592},
  year={2025}
}

@inproceedings{agarwal2024gkd,
  title={On-policy distillation of language models: Learning from self-generated mistakes},
  author={Agarwal, Rishabh and Vieillard, Nino and Zhou, Yongchao and Stanczyk, Piotr and Ramos Garea, Sabela and Geist, Matthieu and Bachem, Olivier},
  booktitle={International Conference on Learning Representations},
  volume={2024},
  pages={21246--21263},
  year={2024}
}

@misc{hunyuan3preview2026,
  author       = {Tencent Hunyuan Team},
  title        = {Hy3 Preview},
  year         = {2026},
  howpublished = {\url{https://github.com/Tencent-Hunyuan/Hy3-preview}},
  note         = {GitHub repository}
}

@article{cobbe2021training,
  title={Training verifiers to solve math word problems},
  author={Cobbe, Karl and Kosaraju, Vineet and Bavarian, Mohammad and Chen, Mark and Jun, Heewoo and Kaiser, Lukasz and Plappert, Matthias and Tworek, Jerry and Hilton, Jacob and Nakano, Reiichiro and others},
  journal={arXiv preprint arXiv:2110.14168},
  year={2021}
}

@article{hendrycks2021measuring,
  title={Measuring mathematical problem solving with the math dataset},
  author={Hendrycks, Dan and Burns, Collin and Kadavath, Saurav and Arora, Akul and Basart, Steven and Tang, Eric and Song, Dawn and Steinhardt, Jacob},
  journal={arXiv preprint arXiv:2103.03874},
  year={2021}
}

@inproceedings{welbl2017crowdsourcing,
  title={Crowdsourcing multiple choice science questions},
  author={Welbl, Johannes and Liu, Nelson F and Gardner, Matt},
  booktitle={Proceedings of the 3rd Workshop on Noisy User-generated Text},
  pages={94--106},
  year={2017}
}

@inproceedings{luong2025towards,
  title={Towards robust mathematical reasoning},
  author={Luong, Minh-Thang and Hwang, Dawsen and Nguyen, Hoang H and Ghiasi, Golnaz and Chervonyi, Yuri and Seo, Insuk and Kim, Junsu and Bingham, Garrett and Lee, Jonathan and Mishra, Swaroop and others},
  booktitle={Proceedings of the 2025 Conference on Empirical Methods in Natural Language Processing},
  pages={35406--35430},
  year={2025}
}

@article{an2025amo,
  title={Amo-bench: Large language models still struggle in high school math competitions},
  author={An, Shengnan and Cai, Xunliang and Cao, Xuezhi and Li, Xiaoyu and Lin, Yehao and Liu, Junlin and Lv, Xinxuan and Ma, Dan and Wang, Xuanlin and Wang, Ziwen and others},
  journal={arXiv preprint arXiv:2510.26768},
  year={2025}
}

@inproceedings{ling-etal-2017-program,
    title = "Program Induction by Rationale Generation: Learning to Solve and Explain Algebraic Word Problems",
    author = "Ling, Wang  and
      Yogatama, Dani  and
      Dyer, Chris  and
      Blunsom, Phil",
    editor = "Barzilay, Regina  and
      Kan, Min-Yen",
    booktitle = "Proceedings of the 55th Annual Meeting of the Association for Computational Linguistics (Volume 1: Long Papers)",
    month = jul,
    year = "2017",
    address = "Vancouver, Canada",
    publisher = "Association for Computational Linguistics",
    url = "https://aclanthology.org/P17-1015/",
    doi = "10.18653/v1/P17-1015",
    pages = "158--167"
}

@article{hendrycks2021apps,
  title={Measuring coding challenge competence with apps},
  author={Hendrycks, Dan and Basart, Steven and Kadavath, Saurav and Mazeika, Mantas and Arora, Akul and Guo, Ethan and Burns, Collin and Puranik, Samir and He, Horace and Song, Dawn and others},
  journal={arXiv preprint arXiv:2105.09938},
  year={2021}
}

@article{wang2023scibench,
  title={Scibench: Evaluating college-level scientific problem-solving abilities of large language models},
  author={Wang, Xiaoxuan and Hu, Ziniu and Lu, Pan and Zhu, Yanqiao and Zhang, Jieyu and Subramaniam, Satyen and Loomba, Arjun R and Zhang, Shichang and Sun, Yizhou and Wang, Wei},
  journal={arXiv preprint arXiv:2307.10635},
  year={2023}
}

@inproceedings{
jain2025livecodebench,
title={LiveCodeBench: Holistic and Contamination Free Evaluation of Large Language Models for Code},
author={Naman Jain and King Han and Alex Gu and Wen-Ding Li and Fanjia Yan and Tianjun Zhang and Sida Wang and Armando Solar-Lezama and Koushik Sen and Ion Stoica},
booktitle={The Thirteenth International Conference on Learning Representations},
year={2025},
url={https://openreview.net/forum?id=chfJJYC3iL}
}

@inproceedings{suzgun2023challenging,
  title={Challenging big-bench tasks and whether chain-of-thought can solve them},
  author={Suzgun, Mirac and Scales, Nathan and Sch{\"a}rli, Nathanael and Gehrmann, Sebastian and Tay, Yi and Chung, Hyung Won and Chowdhery, Aakanksha and Le, Quoc and Chi, Ed H and Zhou, Denny and others},
  booktitle={Findings of the Association for Computational Linguistics: ACL 2023},
  pages={13003--13051},
  year={2023}
}

@misc{aops2025aimei,
  author       = {AIME},
  title        = {2025 AIME I},
  howpublished = {Art of Problem Solving Wiki},
  year         = {2025},
  url          = {https://artofproblemsolving.com/wiki/index.php/2025_AIME_I},
  note         = {Accessed: 2025}
}

@article{lewkowycz2022solving,
  title={Solving quantitative reasoning problems with language models},
  author={Lewkowycz, Aitor and Andreassen, Anders and Dohan, David and Dyer, Ethan and Michalewski, Henryk and Ramasesh, Vinay and Slone, Ambrose and Anil, Cem and Schlag, Imanol and Gutman-Solo, Theo and others},
  journal={Advances in Neural Information Processing Systems},
  volume={35},
  pages={3843--3857},
  year={2022}
}

@misc{rein2023gpqa,
  title        = {GPQA: A Graduate-Level Google-Proof Q\&A Benchmark},
  author       = {David Rein and Betty Li Hou and Asa Cooper Stickland and Jackson Petty and Richard Yuanzhe Pang and Julien Dirani and Julian Michael and Samuel R. Bowman},
  year         = {2023},
  eprint       = {2311.12022},
  archivePrefix = {arXiv},
  primaryClass = {cs.AI}
}

@article{liu2026synlogic,
  title={Synlogic: Synthesizing verifiable reasoning data at scale for learning logical reasoning and beyond},
  author={Liu, Junteng and Fan, Yuanxiang and Zhuo, Jiang and Ding, Han and Hu, Yongyi and Zhang, Chi and Shi, Yiqi and Weng, Shitong and Chen, Aili and Chen, Shiqi and others},
  journal={Advances in Neural Information Processing Systems},
  volume={38},
  pages={100976--100997},
  year={2026}
}

@article{wang2026frontierscience,
  title={FrontierScience: Evaluating AI's Ability to Perform Expert-Level Scientific Tasks},
  author={Wang, Miles and Lin, Robi and Hu, Kat and Jiao, Joy and Chowdhury, Neil and Chang, Ethan and Patwardhan, Tejal},
  journal={arXiv preprint arXiv:2601.21165},
  year={2026}
}

@misc{chou2025autocodebench,
      title={AutoCodeBench: Large Language Models are Automatic Code Benchmark Generators}, 
      author={Jason Chou and Ao Liu and Yuchi Deng and Zhiying Zeng and Tao Zhang and Haotian Zhu and Jianwei Cai and Yue Mao and Chenchen Zhang and Lingyun Tan and Ziyan Xu and Bohui Zhai and Hengyi Liu and Speed Zhu and Wiggin Zhou and Fengzong Lian},
      year={2025},
      eprint={2508.09101},
      archivePrefix={arXiv},
      primaryClass={cs.CL},
      url={https://arxiv.org/abs/2508.09101}, 
}

@article{zhao2025superchem,
  title={SUPERChem: A Multimodal Reasoning Benchmark in Chemistry},
  author={Zhao, Zehua and Huang, Zhixian and Li, Junren and Lin, Siyu and Zhou, Junting and Cao, Fengqi and Zhou, Kun and Ge, Rui and Long, Tingting and Zhu, Yuexiang and Liu, Yan and Zheng, Jie and Wei, Junnian and Zhu, Rong and Zou, Peng and Li, Wenyu and Cheng, Zekai and Ding, Tian and Wang, Yaxuan and Yan, Yizhao and Wei, Tingru and Ming, Haowei and Mao, Weijie and Sun, Chen and Liu, Yiming and Wang, Zichen and Zhang, Zuo and Yang, Tong and Ma, Hao and Gao, Zhen and Pei, Jian},
  journal={arXiv preprint arXiv:2512.01274},
  year={2025}
}

@inproceedings{akyurek-etal-2026-prbench,
    title = "{PRB}ench: Large-Scale Expert Rubrics for Evaluating High-Stakes Professional Reasoning",
    author = {Aky{\"u}rek, Afra Feyza  and
      Gosai, Advait  and
      Zhang, Chen Bo Calvin  and
      Gupta, Vipul  and
      Jeong, Jaehwan  and
      Gunjal, Anisha  and
      Rabbani, Tahseen  and
      Mazzone, Maria  and
      IV, David Randolph  and
      Meymand, Mohammad Mahmoudi  and
      Chattha, Gurshaan  and
      Rodriguez, Paula  and
      Buendia, Diego A. Mares  and
      Singh, Pavit  and
      Liu, Michael  and
      Chawla, Subodh  and
      Cline, Peter  and
      Ogaz, Lucy  and
      Montoya, Ernesto Gabriel Hern{\'a}ndez  and
      Wang, Zihao  and
      Bhatter, Pavi  and
      Ayestaran, Marcos  and
      Liu, Bing  and
      He, Yunzhong},
    editor = "Liakata, Maria  and
      Moreira, Viviane P.  and
      Zhang, Jiajun  and
      Jurgens, David",
    booktitle = "Proceedings of the 64th Annual Meeting of the {A}ssociation for {C}omputational {L}inguistics (Volume 1: Long Papers)",
    month = jul,
    year = "2026",
    address = "San Diego, California, United States",
    publisher = "Association for Computational Linguistics",
    url = "https://aclanthology.org/2026.acl-long.1958/",
    doi = "10.18653/v1/2026.acl-long.1958",
    pages = "42297--42325",
    ISBN = "979-8-89176-390-6"
}

@misc{schulman2020kl,
  author = {John Schulman},
  title = {Approximating KL Divergence},
  year = {2020},
  howpublished = {\url{http://joschu.net/blog/kl-approx.html}}
}

@inproceedings{kazemi-etal-2025-big,
    title = "{BIG}-Bench Extra Hard",
    author = "Kazemi, Mehran  and
      Fatemi, Bahare  and
      Bansal, Hritik  and
      Palowitch, John  and
      Anastasiou, Chrysovalantis  and
      Mehta, Sanket Vaibhav  and
      Jain, Lalit K  and
      Aglietti, Virginia  and
      Jindal, Disha  and
      Chen, Peter  and
      Dikkala, Nishanth  and
      Tyen, Gladys  and
      Liu, Xin  and
      Shalit, Uri  and
      Chiappa, Silvia  and
      Olszewska, Kate  and
      Tay, Yi  and
      Tran, Vinh Q.  and
      Le, Quoc V  and
      Firat, Orhan",
    editor = "Che, Wanxiang  and
      Nabende, Joyce  and
      Shutova, Ekaterina  and
      Pilehvar, Mohammad Taher",
    booktitle = "Proceedings of the 63rd Annual Meeting of the Association for Computational Linguistics (Volume 1: Long Papers)",
    month = jul,
    year = "2025",
    address = "Vienna, Austria",
    publisher = "Association for Computational Linguistics",
    url = "https://aclanthology.org/2025.acl-long.1285/",
    doi = "10.18653/v1/2025.acl-long.1285",
    pages = "26473--26501",
    ISBN = "979-8-89176-251-0"
}
\bibliographystyle{uxbench}

\clearpage
\appendix

\section{System Prompt for Search Agent}

The prompt for search agent specifies the require input arguments and return formats for search tools, as shown in Table \ref{tab:system_prompt}. To remain compatible with continuously evolving search services, the latest tool specifications are dynamically injected into the system prompt at inference time, allowing the agent to interact with up-to-date APIs without modifying the prompt itself.

\begin{table*}[t]
\centering
\caption{System prompt specification of the search agent. The prompt defines the available tools and tool invocation protocol.}
\label{tab:system_prompt}
\small
\begin{tabularx}{\textwidth}{>{\bfseries}p{3.3cm}X}
\toprule

\multicolumn{2}{c}{\textbf{Available Tools}}\\
\midrule

web\_search &
Performs full-text retrieval given one or more search queries and returns relevant textual documents. {Parameter:}
\texttt{\{"query": [str, ...]\}}.
\\[0.8em]

image\_search &
Searches for images using one or more queries and returns image URLs together with metadata (e.g., title and description). {Parameter:}
\texttt{\{"query": [str, ...]\}}.
\\[0.8em]

video\_search &
Performs semantic retrieval over a video corpus using one or more search queries and returns relevant video results. {Parameter:}
\texttt{\{"query": [str, ...]\}}.
\\

\midrule
\multicolumn{2}{c}{\textbf{Tool Invocation Protocol}}\\
\midrule

Invocation Format &
When tool use is required, the model outputs one or more tool calls enclosed by

\texttt{<tool\_calls> ... </tool\_calls>}. Each tool call follows the format

\texttt{<tool\_call>\{tool-name\}<tool\_sep>}

\texttt{<arg\_key>...</arg\_key>}

\texttt{<arg\_value>...</arg\_value>}

\texttt{</tool\_call>}.
\\[1.0em]

Argument Format &
Arguments whose values are lists (e.g., \texttt{query}) must be serialized directly as JSON arrays, e.g., \texttt{["query1", "query2"]}.
\\

\bottomrule
\end{tabularx}
\end{table*}

\end{document}